\documentclass{article}

\PassOptionsToPackage{numbers, compress}{natbib}

\usepackage[preprint]{neurips_2024}

\usepackage{booktabs}
\usepackage{graphicx}
\usepackage{capt-of}
\usepackage[utf8]{inputenc} 
\usepackage[T1]{fontenc}    
\usepackage{hyperref}       
\usepackage{url}            
\usepackage{booktabs}       
\usepackage{amsfonts}       
\usepackage{nicefrac}       
\usepackage{microtype}      
\usepackage{xcolor}         
\usepackage{multirow}
\usepackage{amsmath}
\usepackage{subcaption}

\title{Transitive Vision-Language Prompt Learning for Domain Generalization}

\author{%
  Liyuan Wang$^{1}$, Yan Jin$^{1}$, Zhen Chen$^{3}$, Jinlin Wu$^{4}$, Mengke Li$^{2}$, Yang Lu$^{1}$\thanks{Corresponding Author: Yang Lu (luyang@xmu.edu.cn)}, Hanzi Wang$^{1}$\\
  $^1$Fujian Key Laboratory of Sensing and Computing for Smart City, \\
  School of Informatics Xiamen University, Xiamen, China\\
  $^2$Guangdong Laboratory of Artificial Intelligence and Digital Economy (SZ), Shenzhen, China\\
  $^3$Centre for Artificial Intelligence and Robotics, Hong Kong Institute\\ of Science Innovation Chinese Academy of Sciences\\
  $^4$Institute of Automation, Chinese Academy of Sciences, Beijing, China\\
  \texttt{lywang0802@163.com}, \texttt{jinyan7973@gmail.com}, \texttt{zhen.chen@cair-cas.org.hk} \\
  \texttt{jinlin.wu@nlpr.ia.ac.cn}, \texttt{limengke@gml.ac.cn}, \texttt\{luyang, hanzi.wang\}@xmu.edu.cn \\
}

\begin{document}
\maketitle
\begin{abstract}
  The vision-language pre-training has enabled deep models to make a huge step forward in generalizing across unseen domains. 
  The recent learning method based on the vision-language pre-training model is a great tool for domain generalization and can solve this problem to a large extent. 
  However, there are still some issues that an advancement still suffers from trading-off between domain invariance and class separability, which are crucial in current DG problems.
  In this paper, we introduce a novel prompt learning strategy that leverages deep vision prompts to address domain invariance while utilizing language prompts to ensure class separability, coupled with adaptive weighting mechanisms to balance domain invariance and class separability. 
  Extensive experiments demonstrate that deep vision prompts effectively extract domain-invariant features, significantly improving the generalization ability of deep models and achieving state-of-the-art performance on three datasets.
\end{abstract}

\section{Introduction}
\label{Introduction}
In real-world applications, visual data often presents itself in various domain forms, reflecting the vast diversity and unpredictability of our environment. 
In this context, the task of domain generalization (DG) plays an important role in building models that perform both robustly and consistently in variable and unseen environments. 
It aims to investigate how to efficiently generalize models trained in specific or multiple source domains to completely new, unseen target domains~\cite{blanchard2021domain}. 
This challenge is particularly salient in sectors with highly variable data distributions, such as medical imaging, autonomous driving, and natural language processing.

\begin{figure*}[thbp]
    \centering
    \vspace{-10mm}
	\includegraphics[width=0.9\textwidth]{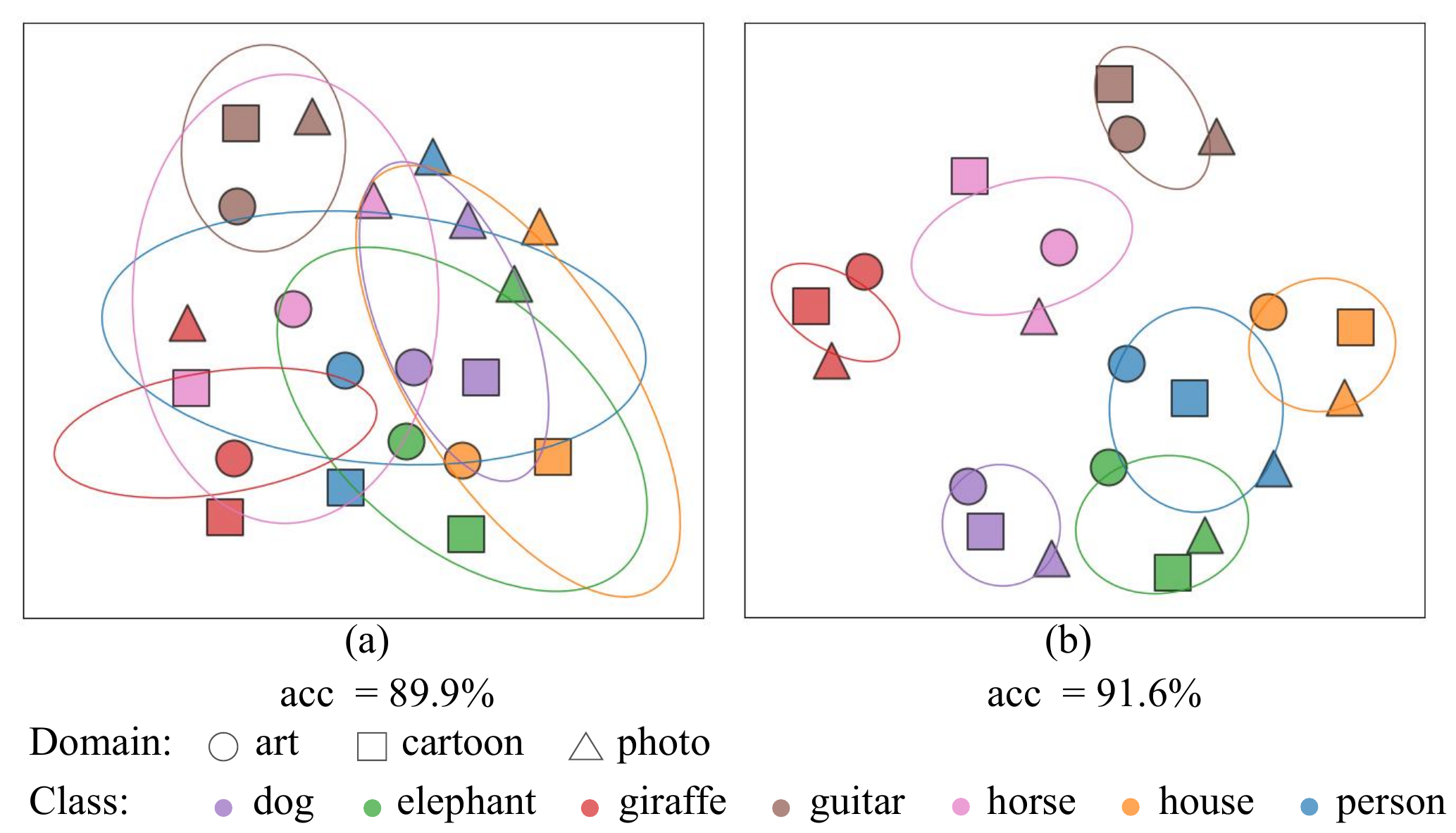}
	\caption{2D feature space visualization on the PACS dataset by (a) jointly training vision and language prompts and (b) our proposed learning strategy. Shapes represent domains, and colors denote classes. The ellipses illustrate the degree of domain invariance by fitting a Gaussian model.}  
	\label{MDS}
\end{figure*}

In the field of DG, various techniques are employed to enhance the generalization capabilities of models to unseen domains. 
These methods include data augmentation~\cite{shi2020towards}, feature disentanglement~\cite{li2017deeper}, meta-learning~\cite{li2019feature}, and invariant risk minimization~\cite{zhang2017mixup}. 
These approaches essentially revolve around two core strategies. 
One is the learning of domain-invariant features~\cite{lu2022domain,shao2019multi,li2018domain}, and the other area emphasizes exploiting domain-specific information~\cite{liang2023comprehensive,bui2021exploiting}. 
These methods generally use traditional vision models. 
The main problem with these approaches is to find a balance between domain invariance and class specificity. Data imbalances among domains~\cite{yang2022multi} make this balance difficult. 
Models are required that can handle these changes without compromising generalization capabilities.

However, there exists a notable drawback of traditional vision models in DG, that is they rely only on image features~\cite{zheng2022prompt}. To address this limitation, CLIP~\cite{radford2021learning} is introduced in DG by combining image and text data through a contrastive learning framework.
The language component can add information about each domain, benefiting model classification and generalization to unseen domains. 
This pre-trained model leverages the strengths of both image and text features, enhancing the model's ability to generalize across diverse domains effectively.
Fully fine-tuning CLIP on the task of DG may result in overfitting to specific features of the training data, thereby losing the broad generalization capabilities necessary for DG~\cite{kumar2022fine}.
Existing solutions adopting CLIP for DG often involve modifying language or vision components separately~\cite{zhang2023domain,shu2023clipood}, in order to adapt the CLIP model to the possibility of domain shifts on unknown test data. 
Only modifying the vision component while keeping the textual part unchanged tends to make the image features converge~\cite{dong2022lpt}, focusing on learning domain invariant knowledge. 
On the other hand, altering only the language component can lead to a divergence in the representation~\cite{zhang2023domain}, failing to obtain specific information required for accurate categorization.
However, these approaches may disrupt CLIP's vision-language alignment between image and text features.

To solve the misalignment of language vision modules in existing methods, we propose a joint learning strategy. 
Jointly training vision and language prompts can lead to excessive fluctuation in the feature space, impeding the learning process.
Figure~\ref{MDS} shows the feature visualization map using MDS downscaling on the PACS training set where the target domain is the sketch domain and compares the effect of jointly training vision and language prompts with our proposed learning strategy.
As shown in Figure~\ref{MDS}(a), there is a significant overlap in the confidence ellipse of each category after applying the joint learning method, which indicates that the boundaries and domain invariance between categories in the feature space are not clear enough. 
A possible reason is that the text features tend to diverge when the image features start to cluster. 
This divergence in the text features inhibits further clustering of the image features, which is problematic for the learning process.
To fully utilize the integrated visual and textual information while conquering the problem of joint learning, we propose the Transitive vision-language Prompt Learning (TPL) framework in this paper. 
It balances the feature space dynamically between vision and language components during the learning process. 
To circumvent this issue, the proposed TPL begins by first adapting the vision component with prompts and subsequently adjusting the language component. 
It strategically uses both vision and language prompts to adapt the model to different domains without breaking the CLIP image-text alignment. 
This transitive Vision-Language method ensures a more stable and effective learning process, maintaining the alignment between image and text representations. 
It is beneficial to obtain domain-specific knowledge without compromising the model's ability to generalize.
The feature space produced by the proposed TPL is shown in Figure~\ref{MDS}(b). 
The visualization shows tight clustering of the same classes across domains, indicating strong domain invariance.
Furthermore, there is a pronounced separation between different classes within the same domain, underscoring the ability to maintain class separability. 
In contrast, Figure~\ref{MDS}(a) cannot learn either domain invariance or class separability.

The main contributions of this work are as follows:
\begin{itemize}
    \item We propose a new learning framework that tunes both vision and language prompts of a pre-trained CLIP model, addressing the complexity of balancing image-text alignment for the task of DG.
    \item We explore the necessity of a proper learning strategy to maintain the image-text alignment within the CLIP model and proposed TPL, which keeps the balance between domain invariance and class separability.
    \item The proposed TPL achieves state-of-the-art results on three DG benchmarks, emphasizing the importance of transitive learning towards different modalities in improving DG performance.
\end{itemize}

\section{Related Work}

\subsection{Domain Generalization}
DG research aims to improve the performance of models in unseen domains. 
The methods for DG primarily focus on learning domain-invariant features or leveraging domain-specific knowledge. 
Feature-based methods aim to learn representations that are invariant across domains. 
Methods like invariant risk minimization (IRM)~\cite{finn2017model} and domain adversarial neural networks (DANN) are prominent. 
IRM focuses on learning features that yield invariant predictors across domains, while DANN~\cite{ganin2016domain} uses a domain classifier to encourage the model to learn features that are indistinguishable across different domains. 
Meta-learning has been effectively utilized in DG to simulate domain shifts during training. 
Model-Agnostic Meta-Learning (MAML)~\cite{finn2017model} optimize models to quickly adapt to new domains with minimal fine-tuning. 
This approach trains models on a variety of tasks to enhance their generalization ability. 
Data augmentation methods like Mixup~\cite{zhang2017mixup} and CutMix~\cite{yun2019cutmix} generate synthetic training examples by combining features and labels from different domains, enhancing the generalization ability of models. 
Ensemble methods~\cite{khattak2023self} combine predictions from multiple models or model variants, improving robustness against domain shifts.

\subsection{Vision-Language Pre-training}

The use of pre-trained models in DG especially from large-scale datasets has shown promise. 
These models are initially trained on different data and can capture a wide range of features. 
Vision-language pre-training (VLP) mainly learns the semantic correspondence between different modalities by pre-training on large-scale data~\cite{chen2023vlp}. 
VLP began with models like ViLBERT~\cite{lu2019vilbert} and LXMERT~\cite{tan2019lxmert}, which used dual-stream architectures to process visual and textual information separately. 
It combines images with natural language descriptions, enabling models to perform a variety of tasks such as zero-shot learning, object recognition, etc., with minimal task-specific fine-tuning. 
It is particularly effective in addressing the challenges associated with the DG problem. CLIPood ~\cite{shu2023clipood} focuses on enhancing CLIP's ability to generalize to OOD scenarios. 
It leverages the robust image-text representations of CLIP and fine-tunes them on a diverse set of data to improve performance on unseen domains. 
However, CLIPood is limited by its reliance on extensive fine-tuning. 
DPLCLIP~\cite{zhang2023domain} involves using domain-specific prompts to adapt CLIP to new domains efficiently. 
While it offers a more computationally efficient way to adapt CLIP, this method may still struggle with domains that are significantly different from the training data, as the reliance on domain-specific prompts can limit the model's flexibility. 
By utilizing CLIP's pre-trained abilities, we aim to address the challenges of DG, where models need to perform accurately across various unseen domains. 
Our approach involves adapting CLIP to diverse settings, demonstrating its effectiveness in handling domain shifts and data variability.

\subsection{Prompt Learning}

Originating in natural language processing (NLP), prompt learning customizes inputs to steer pre-trained models towards new tasks. This approach has expanded into computer vision with Vision Prompt Tuning (VPT), which adapts existing models to new tasks or domains by adjusting a small subset of parameters.
CLIP uses language prompts to effectively classify images across various domains, demonstrating the versatility of prompt learning in multimodal contexts. MaPLe~\cite{khattak2023maple} proposes joint learning for vision and language prompts, but modifies both modalities jointly. 
The joint training approach may not only skew the model's understanding by favoring one modality but also underutilize the unique strengths inherent to each modality, leading to challenges in achieving a balanced and effective learning process.  
Although CoOp~\cite{zhou2022learning}, CoCoOp~\cite{zhou2022conditional}, and MaPLe represent significant strides in prompt learning, their approaches exhibit limitations, particularly in multimodal learning contexts. 
These include a focus on single-modality learning, static or inflexible prompt structures, and challenges in effectively integrating and balancing multiple modalities within the learning process.
In this work, the proposed TPL differs from existing methods as we utilize multi-modal learning to balance domain invariance and class separability in the DG field. 
We discovered that vision prompts effectively achieve domain invariance, while language prompts enhance class separability.
Additionally, we leverage the coupling characteristics of modalities to generate domain-specific language prompts for each domain, using the domain-invariant image features learned through a generator.

\section{Preliminaries}
\paragraph{Domain Generalization.} 
In the setting of DG, it is usually assumed that there are $M$ source domains, denoted as $ D_1, D_2, \ldots, D_M $.
Each domain $ D_i $ consists of a dataset $ \{ (x_i^m, y_i^m) \}_{i=1}^{N_m} $, where $ x_i^{m} $ represents the $i$-th sample in the $m$-th domain, and $y_i^{m}$ is its corresponding label. $N_m$ is the total number of samples in the $m$-th domain.
The goal of DG is to ensure that a model trained on the source domains performs well on the target domain $D_{M+1}$, which is not available during training. 
A model $f_\theta$ with parameters $\theta$ is trained on the union of all datasets from the source domains. 
The training objective is often to minimize the empirical risk $ R $ on the source domains~\cite{ben2010theory}:
\begin{equation}
	\begin{aligned}
		 R  = \min_\theta \sum_{m=1}^M \sum_{i=1}^{N_m} l \left( f_\theta \left(x_i^{m}\right), y_i^{m} \right),
	\end{aligned}
\end{equation}
where $l$ is a loss function measuring the prediction error of the model $f_\theta$.
The performance of the model is evaluated on an unseen target domain. The key challenge is that the data distribution in $ D_{M+1} $ may be different from those in the source domains $ D_1, D_2, \ldots, D_M $. The generalization ability of the model is quantified by its performance on $ D_{M+1} $, typically measured by accuracy or other relevant metrics in classification tasks.
The essence of DG is to learn a model that captures domain-invariant features~\cite{johansson2019support}, which are effective across both the seen source domains and the unseen target domain. This requires the model to not only perform well on the training data but also to generalize to new, unseen data distributions.

\paragraph{Contrastive Language-Image Pre-training (CLIP).} 
CLIP is trained on a large dataset of images and their corresponding text captions to learn visual representations that can be effectively paired with textual representations.  
CLIP is to align the image and text embeddings in a common multi-modal space where semantically similar concepts, regardless of modality, are closer to each other. 
Drawing on the principles of CLIP's zero-shot learning, language prompts are tailored for each category $c$, crafting descriptors like ``a photo of a [\texttt{CLASS}]'', where [\texttt{CLASS}] is substituted with the actual category name, providing a unique linguistic context for each class. 
CLIP comprises two main components, the image encoder $E_I(\cdot)$ and the text encoder $E_L(\cdot)$, which are responsible for transforming raw images and text into embeddings, respectively. 
CLIP employs a contrastive loss function, which is designed to minimize the distance between the correct image-text pairs and maximize the distance between mismatched pairs.

\section{Proposed Method}

\begin{figure*}[thbp]
    \centering
    \vspace{-10mm}
	\includegraphics[width=0.9\textwidth]{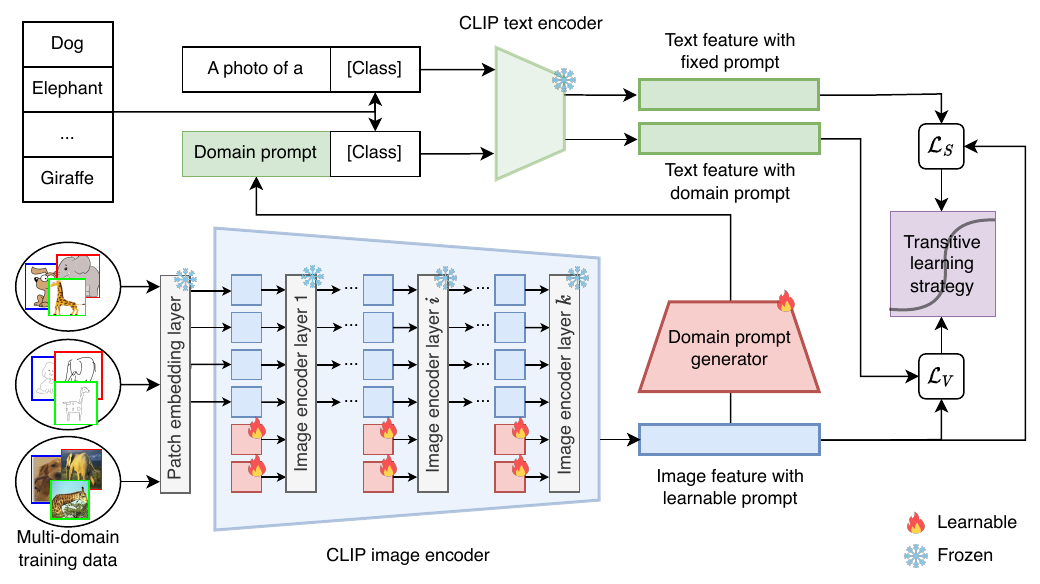}
	\caption{Overview of the proposed TPL. Vision prompts are trained to enhance domain invariant features. Then they are used as the input of the domain prompt generator to produce domain-specific prompts for the text encoder. Finally, the transitive learning strategy combines the above two components to balance domain invariance and class separability during the whole tuning process.}  
	\label{fig2} 
\end{figure*}

In this paper, we propose the Transitive vision-language Prompt Learning (TPL) framework to address the problem of DG from the perspective of strategically jointly learning vision and language prompts.
In TPL, we have two components and acknowledge the significance of both vision and language prompts.
The vision component guarantees domain invariance across features from different domains, while the language component ensures class separability during the learning process.
The produced features with domain invariance with class separability are produced by a newly proposed transitive learning strategy for joint learning. 
The proposed framework is illustrated in Figure~\ref{fig2}. 

\subsection{Domain-Invariant Vision Prompt Learning}
\label{vision}

The first component is primarily a learnable vision prompt, which is plugged in to learn domain invariant features. 
Building on the CLIP architecture, the model's ability to extract and generalize key characteristics across different domains can be enhanced via leveraging deep vision prompts, thereby improving its robustness and adaptability in various unseen environments~\cite{khattak2023maple,khattak2023self,zhu2023prompt}.

Specifically, this model processes the input image into a sequence of patch embedding, effectively capturing the essential features of the image for DG tasks.  
A series of learnable tokens are introduced within the vision component. 
In each layer, the image tokens and class tokens are processed through the transformer blocks, denoted as $ \mathcal{V}_k $, for $k = 1, 2, \ldots, K$. The learnable tokens are denoted as $\mathbf{p}_1, \mathbf{p}_2, \ldots, \mathbf{p}_K$. 
The final image representation is obtained by projecting the class token of the last transformer layer through an embedding space via image projection, culminating in an embedding that captures the essence of the image guided by the deep vision prompts. 
\begin{equation}
	\begin{aligned}
		 [\mathbf{c}_k, \mathbf{e}_k(x)] &= \mathcal{V}_k([\mathbf{c}_{k-1}, \mathbf{e}_{k-1}(x), \mathbf{p}_{k-1}]),
	\end{aligned}
\end{equation}
for $k=1, 2, \ldots,K$. 
Note that $\mathbf{e}_k(x)$ represents the embeddings of the image $x$ at layer $k$, while $\mathbf{e}_0(x)$ represents the initial set of patch embeddings obtained from the input image $x$.
It is a combination of the class token $\mathbf{c}_k$ and the embeddings $\mathbf{e}_k(x)$ at the $k$-th layer.
$\mathbf{c}_{k-1}$ represents the class token from the previous layer $k-1$. $\mathbf{p}_{k-1}$ represents the prompt embeddings added at the  $ (k-1) $-th layer to guide the transformer's focus. The transformer block is denoted as $\mathcal{V}_k$, which takes the class token, patch embeddings, and prompt embeddings from the previous layer and processes them to produce the output for the  $k$-th layer.

Such kind of prompting enhances the flexibility of the CLIP model, allowing pre-trained models to adjust to different feature hierarchies and forcing prompts to fit towards acquiring domain-invariant knowledge. 
This advanced prompting~\cite{khattak2023maple} allows the CLIP model to foster a deeper and more hierarchical understanding of image features. 
Given a training sample $x_i^m$ in the $m$-th domain with its corresponding crafted class descriptor $y_i$ (``a photo of a [\texttt{CLASS}]''), we can obtain the domain-invariant image feature $I_i^m$ and the domain-agnostic text feature $T_{y_i}$:
\begin{align}
    I_i^m = \mathbf{e}_K(x_i^m),~~~T_{y_i} = E_L(y_i).
\end{align}
The loss function is given by:
\begin{equation}
	\begin{aligned}
		 \mathcal{L}_{V} = -\sum_{m=1}^M\sum_{i=1}^{N_m} \log \frac{\exp(\text{sim}(I_i^m, T_{y_i}) / \tau)}{\sum_{j=1}^{C} \exp(\text{sim}(I_i^m, T_j) / \tau)},
    \label{eq:LV}
	\end{aligned}
\end{equation}
where $\text{sim}(\cdot, \cdot)$ denotes the similarity between text feature and the corresponding image feature.
$\tau$ is a temperature parameter that scales the logits. 
This loss represents the sum of the losses of the $M$ domains and $N_m$ training samples in the $m$-th domain.

\subsection{Class-Aware Language Prompt Learning}
\label{language}
The second component is primarily a learnable language prompt generator.
Utilizing only vision prompts tends to focus on domain-invariant features while neglecting the diversity of textual information in CLIP that can further enhance DG by improving class separability. 
Leveraging the stability of textual representation, which inherently lacks domain-specific differentiation, the proposed TPL maintains uniform language prompts across all domains. 
Each class's text embedding clusters together, aligning with the domain-agnostic language context. 
This homogeneity in text embeddings ensures that while image features are drawn towards domain-generic clusters through deep prompts, the text remains a constant reference point. 
This alignment fosters concentrated learning of domain invariance in image features while retaining the crispness of textual discrimination. 
Our proposed TPL employs a prompt generator ${G}(\cdot)$ that receives domain-invariant image features from each domain to generate domain-specific language prompts. 
We can obtain the domain-specific language prompt $v^m$ and the domain-specific text feature $\overline{T}_i^m$ in the $ m $-th domain:
\begin{align}
v^m &= \frac{1}{N_m}\sum_{i=1}^{N_m}{G}({I_i^m}),\\
\overline{T}_i^m &= E_L([v^m ,y_i]),
\end{align}
This generator adaptively produces domain-specific language prompts by averaging the generated prompts from all the image features in the $m$-th domain.
It captures both shared and unique image attributes. Such synergistic text-image interplay~\cite{khattak2023maple} enhances the model's learning capacity, optimizing its generalizability across new domains.
This generator is comprised of a three-layer MLP, specifically designed to respond to the unique demands of each domain. The learnable language prompts, combined with the individual sample labels denoted by [\texttt{CLASS}], form a new textual context. This adaptive mechanism ensures that the prompts are not static but evolve with the domain-specific characteristics of the dataset, allowing for more dynamic and flexible integration of visual and textual information. 

We employ a cross-entropy loss that intertwines the deep vision prompts with language prompts generated by the generator. The deep image prompts are designed to enrich the image features with domain-agnostic information, while the language prompts are adaptively generated to provide class-specific language prompts. The loss function is given by:
\begin{align}
\mathcal{L}_{S} = -\sum_{m=1}^M\sum_{i=1}^{N_m} \log \frac{\exp(\text{sim}(I_i^m, \overline{T}_{y_i}^m) / \tau)}{\sum_{j=1}^{C} \exp(\text{sim}(I_i^m, \overline{T}_j^m) / \tau)}.
\label{eq:LS}
\end{align}%

\subsection{Adaptive Fusion for Domain Robustness}
To effectively generalize across different domains, it is essential to balance the stability of the original CLIP model's semantic knowledge and the flexibility introduced by domain-specific prompts. 
We ensure through feature adaptive fusion that the model not only preserves the rich semantic knowledge embedded in the original CLIP model but also benefits from the adaptability introduced by the domain-specific prompts. 
We dynamically fuse image features with the original image features from a pre-trained and frozen CLIP model. 
Similarly, the text features generated by the prompt generator are adaptively merged with the original text features. 
The model retains the original semantic knowledge while incorporating domain-specific language knowledge. 
To enhance the learning of domain-invariant features, we further adaptively mix the domain-specific prompted text features with the domain-agnostic text features. 
This feature fusion is designed to balance the domain-specific and general features, enabling the model to generalize better across different domains. 
The features are fused as follows:
\begin{align}
   {I^{m}}' &= I^{m} + \mathbf{P} \circ I^{m}_{\text{orig}}, \\
   {\overline{T}^{m}}' &= \overline{T}^m + \mathbf{Q} \circ T,  \\
   T' &= T + \mathbf{R} \circ \overline{T}^m,
\end{align}%
where $ \mathbf{P} $, $ \mathbf{Q} $, $  \mathbf{R} $ are all learnable parameters. $ I^{m}_{\text{orig}} $ denotes the image features of the $m$-th domain after the original model clip.
The loss functions in Eq.~\ref{eq:LV} and Eq.~\ref{eq:LS} use fused features.

\subsection{Transitive Learning Strategy}

As shown in Figure~\ref{MDS}, we discerned a pivotal challenge when utilizing both vision and language prompts for DG.
Different from the task of few-shot learning~\cite{zhang2021tip} and incremental learning~\cite{wang2023attriclip}, jointly training vision and language prompts may lead to the problem of balancing domain invariance and class separability in DG because balancing domain-specific and domain-invariant features can be challenging. 
To address this problem, we propose a transfer learning strategy to adaptively obtain domain invariance and class separability by merging the domain-invariant vision prompt learning loss with the class-aware language Prompt Learning loss.
The key to this strategy is to dynamically adjust the weights based on the inter-domain distances within the source domains. 
The weight coefficient is computed as:
\begin{align}
\lambda^{(t)} = -\ln\left(\frac{d^{(t)} (T - t)}{\theta}\right),
\end{align}%
where $d^{(t)}$ is the average inter-domain distance in the source domain, $T$ is the total number of iterations, $t$ is the current iteration, and $\theta$ is a scaling factor. 
Therefore, $\lambda$ measures domain invariance that scales the inter-domain average aggregation of the training set to an appropriate range, which is updated at each checkpoint.
The weights for domain invariance $ w_{V}^{(t)}$ and class separability $ w_{S}^{(t)} $ are then given by:
\begin{align}
w_{V}^{(t)} = \frac{1}{1 + \lambda^{(t)}}, \quad w_{S}^{(t)} = \frac{\lambda^{(t)}}{1 + \lambda^{(t)}}.
\end{align}
When the initial distance $ d = 1 $, which implies the maximum distance, $ \lambda $ becomes 0; When the distance $ d $ approaches 0, which implies the minimum distance, $ \lambda $ tends towards infinity. Notably, as $ \lambda $ increases, the weight allocated to domain invariance diminishes while that for class separability escalates.  
The total final loss on epoch $t$ is as follows:
\begin{align}
\mathcal{L}^{(t)}_{TLS} = w_{V}^{(t)} \cdot \mathcal{L}_V + w_{S}^{(t)} \cdot \mathcal{L}_S.
\end{align}
This adaptive approach ensures that each training iteration is uniquely tailored based on the characteristics of the source domain, thereby fine-tuning the equilibrium between domain invariance and class separability. Such dynamic balancing is crucial for the model's effective learning across varied domains, as it adeptly manages the interplay between domain-invariant and domain-specific features.

In Sec~\ref{Ablation study}, we demonstrated that, throughout the training process, the inter-domain distances progressively decrease, along with a corresponding reduction in domain invariance weights. Additionally, we adaptively generated weights for each domain, based on the varying inter-domain distances specific to each domain.
\section{Experiments}

\subsection{Experimental Setting}

\paragraph{Datasets.} We conduct our evaluations on three benchmarks: PACS, VLCS and OfficeHome. 
PACS~\cite{li2017deeper} consists of four domains, including photo, art, cartoon, and sketch, where the challenges lie in their significant stylistic differences for stringent generalization testing. 
VLCS~\cite{fang2013unbiased} aggregates images from various sources like VOC2007, LabelMe, Caltech101, and SUN09, contributing to various object representations and scene contexts. 
OfficeHome~\cite{venkateswara2017deep} provides an office-centric collection with domains such as art, clipart, product, and real-world, showcasing everyday objects in office environments and is well-suited for practical domain adaptation tasks. 

\paragraph{Implementation Details.} We conduct our experiments based on the DomainBed~\cite{gulrajani2020search} implementation. we use \text{ViT-B/16}  model~\cite{dosovitskiy2020image} pretrained on ImageNet for all experiments.  
We adopt the training-domain validation set method~\cite{gulrajani2020search} for model selection.
Specifically, each source domain is split into training and validation subsets. 
We use the AdamW~\cite{loshchilov2017decoupled} optimizer with the cosine learning rate strategy and the same prompt ensemble as promptSRC~\cite{khattak2023self}. 
We adopt a learning rate of $ 3 \times 10^{-5} $, set $ \theta=1.25 \times 10^{3} $ and train for 5000 iterations. 
The final evaluation results are averaged with those from the original pre-trained model to preserve the rich semantic capabilities of the original CLIP model. 
We report the average result and the standard deviation of three runs with random seeds. 

\subsection{Comparative Results}

We report the average precision of the test domains for the three datasets as shown in Table \ref{tab:comparison_methods}. 
We compared fine-tuning the model using techniques such as ERM, CORAL~\cite{sun2016deep} and DANN~\cite{ganin2016domain}.
We compare the proposed TPL against the state-of-the-art methods utilizing the CLIP pre-trained model for DG, including MIRO~\cite{cha2022domain}, DPL, and CLIPood. 
Because of the exactly same experimental protocol, the results from the original publications of these methods are summarized in Table \ref{tab:comparison_methods}. 
We achieve the best performance on each of the three benchmark datasets.
\begin{table}[t]
	\centering
	\caption{Comparison of different methods on benchmark datasets. The average accuracy of the three benchmarks is also reported. The best results are shown in \textbf{bold}, and the second-best results are shown in \underline{underline}.}
    \label{tab:comparison_methods}
	\begin{tabular}{ c c c c c c}
		\toprule
        Methods & Type & PACS & VLCS & OfficeHome & Avg. \\
		\midrule
		ERM &  Fine-tuning & 92.9 & 82.7 & 78.1 & 84.6 \\
        CORAL~\cite{sun2016deep} &  Fine-tuning & 93.2 & 82.0 & 78.9 & 84.7 \\
        DANN~\cite{ganin2016domain} &  Fine-tuning & 93.8 & 83.2 & 78.8 & 85.3 \\
        CLIP~\cite{radford2021learning} & Zero-shot & 95.8 & 76.6 & 79.9 & 84.1 \\
        MIRO~\cite{cha2022domain} & Fine-tuning & 95.6 & 82.2 & 82.5 & 86.8 \\
        DPL~\cite{zhang2023domain} & Prompt learning & 97.3 & 84.3 & 84.2 & 88.6 \\
        CLIPOOD~\cite{shu2023clipood} & Fine-tuning & \underline{97.3} & \underline{85.0} & \underline{87.0} & \underline{89.7}\\
        TPL  & Prompt learning & {\bf97.4 }& {\bf85.4} & {\bf87.5} & {\bf90.1} \\
		\bottomrule
	\end{tabular}
\end{table}

\begin{figure*}[!t]
  \footnotesize
  \centering
  \begin{tabular}{cc}
    \includegraphics[width=0.45\linewidth]{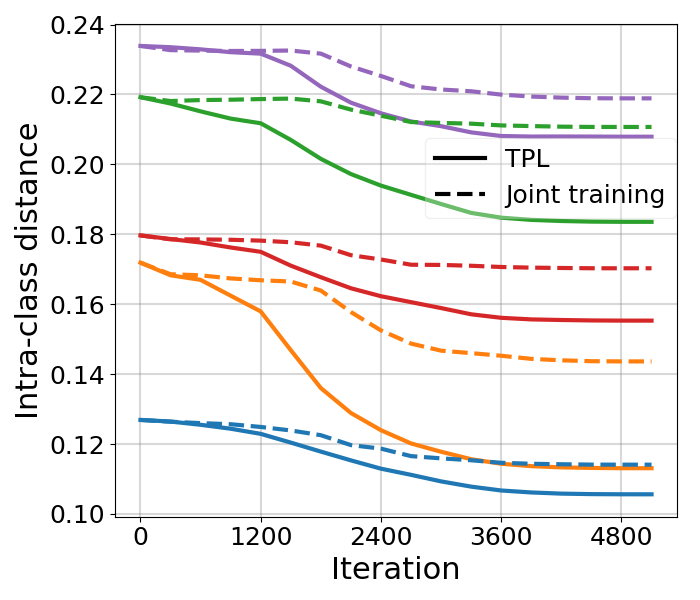} &
    \includegraphics[width=0.45\linewidth]{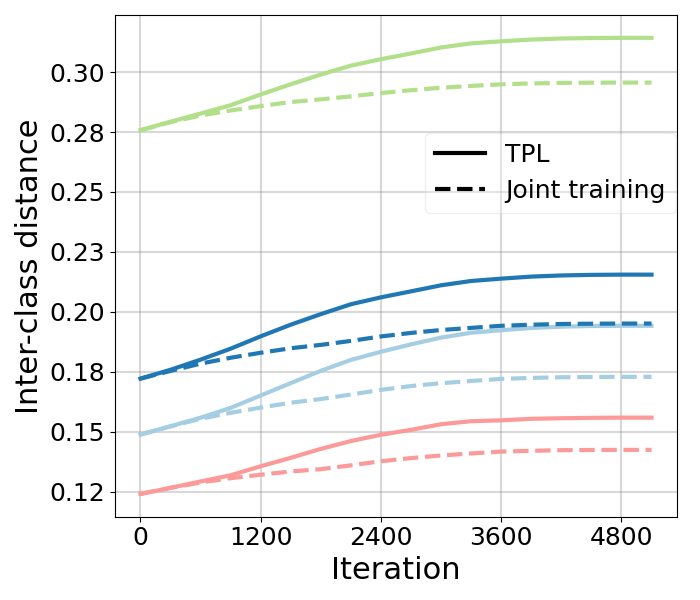} \\
    (a) Domain invariance & (b) Class separability \\
  \end{tabular}
  \caption{(a) The average distance between any two domains in each class on the VLCS dataset demonstrates inter-class domain invariance. Different colors indicate different classes. (b) The average distance between any two classes in each domain on the PACS dataset demonstrates intra-domain class separability. Different colors indicate different domains.}
  \label{Domain invariance and Class separability}
\end{figure*}

\subsection{Domain Invariance and Class Separability}
Balancing domain invariance and class separability during joint prompt learning is the key to the proposed TPL.
It is crucial to ensure that our model adapts to diverse domains and maintains clear class distinctions.
Therefore, we specifically design a set of experiments to validate the capability of the proposed TPL to strike a balance between domain invariance and class separability. 

\paragraph{Enhanced Domain Invariance.} 
As shown in Figure \ref{Domain invariance and Class separability}(a), we show the average distance between any two domains in each class on the VLCS dataset. 
The smaller distance indicates more similarity between different domains with the same class in the feature space. 
Compared with the joint training image and text prompts, TPL is shown to be more effective in reducing the distance across different domains.
It confirms the importance of domain invariance for DG, essential for models expected to perform well in unseen environments.

\paragraph{Improved Class Separability.} 
As shown in Figure \ref{Domain invariance and Class separability}(b), we show the average distance between any two classes in each domain on the PACS dataset. 
The larger distance indicates better separability between the classes within the same domain in the feature space. 
Compared with the joint training image and text prompts, our approach greatly increases the inter-class distance within the same domain. 
It enhances the model's generalization ability to distinguish among different classes.
Thus, it validates that TPL can learn not only better domain invariance but also better classification ability. 
\label{Ablation study} 
\subsection{Ablation Study}

\begin{table}[t]
    \centering
    \caption{Comparing different prompt designs on the VLCS dataset.}
    \label{tab2}
    \begin{tabular}{ c c c c c c}
        \toprule
        Method                 & Accuracy           \\
        \midrule
        CILP     & 76.6               \\
        CILP w/ language prompt     & 82.5               \\
        CILP w/ vision prompt       & 83.9               \\
        CILP w/ joint training w/o Adaptive Fusion                & 84.5               \\
        CILP w/ joint training                & \underline{84.7}              \\
        TPL                    & {\bf85.4}                  \\
        \bottomrule
    \end{tabular}
\end{table}

\begin{table}[t]
    \centering
    \caption{Comparison of different learning strategies on OfficeHome dataset.}
    \label{tab3}
    \begin{tabular}{ c c c c c c}
        \toprule
        Strategy                 & Accuracy          \\
        \midrule
        Baseline                &      85.0          \\
        Alternate learning                &      85.5          \\
        Two stage~\cite{kang2019decoupling}                &      85.8          \\
        Cumulative learning ~\cite{zhou2020bbn}            &      \underline{86.7}         \\
        TPL                                            &     {\bf87.5}        \\
        \bottomrule
    \end{tabular}
\end{table}

In the ablation study, we explore several key aspects to validate the effectiveness of our proposed methodologies. 
We first demonstrate the effectiveness of the method by comparing different prompt designs. 
It validates the superiority of the proposed prompt learning scheme.
Then, the effectiveness of adaptive weighting is examined, showcasing the proposed TPL's exceptional dynamic weight balancing during training.

\paragraph{Effectiveness of the Prompt Design.} 
We first evaluate the performance of different possible prompting design choices as ablation for our proposed prompting. 
We demonstrate superior results when compared to the strategies of adding prompts only to images, only to text, or to both jointly. 
In Table \ref{tab2}, we present the results on the VLCS dataset. 
Vision prompting shows improvements over language prompting, indicating that prompts learned at the deep vision component provide better domain invariance. 
Although combining the above two methods alone further improves the performance, it does not learn both domain invariance and class separability as well, compared to the proposed TPL .

\begin{figure*}[!t]
  \footnotesize
  \centering
  \begin{tabular}{cc}
    \includegraphics[width=0.45\linewidth]{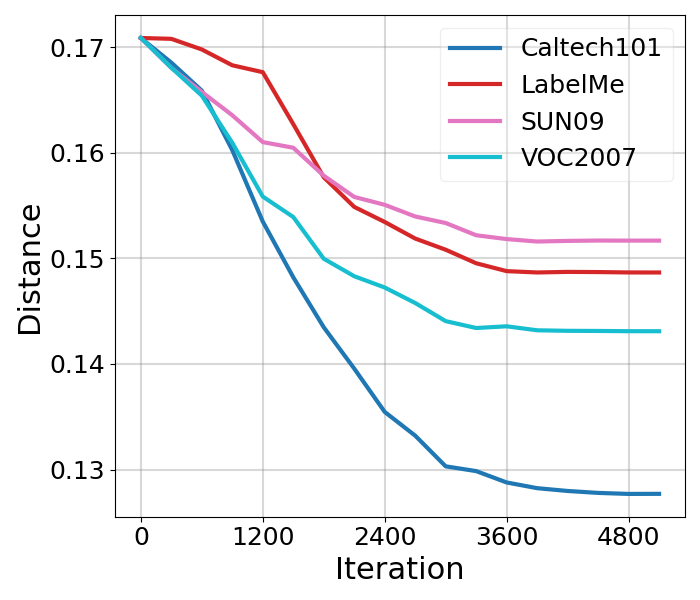} &
    \includegraphics[width=0.45\linewidth]{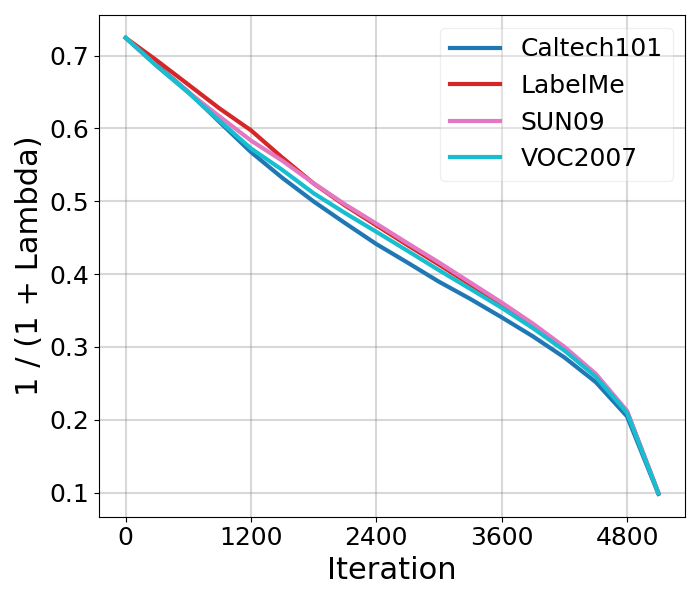} \\
    (a) Inter-domain distance & (b) Domain invariant weights \\
  \end{tabular}
  \caption{(a) Average inter-domain distance on the VLCS dataset. (b) Domain invariant weights on the VLCS dataset.}
  \label{Adaptive Weighting}
\end{figure*}

\paragraph{Effectiveness of Adaptive Weighting.} 
The key to the effectiveness of TPL is the transitive learning scheme that dynamically optimizes vision and language prompts to achieve features with domain invariance and class separability.
As illustrated in Figure \ref{Adaptive Weighting}(a), during the training process, the average inter-domain distance across each domain gradually decreases, leading to an increasing similarity in the feature space between domains. 
Figure \ref{Adaptive Weighting}(b) demonstrates that, after our adaptive weighting transformation, even though the inter-domain distances vary across each domain, the domain invariance weights can be scaled between 0 and 1, progressively decreasing from 1 to 0. 
Adaptive weighting based on the average distance between domains dynamically balances the weights of domain invariance and class separability. 

In addition to the proposed transitive learning scheme, there are also some strategies to dynamically learn from two branches.
Therefore, we ablate our strategy with others to show the its superiority in the context of DG.
As shown in Table \ref{tab3}, compared to BBN~\cite{zhou2020bbn} and two-stage decoupling~\cite{kang2019decoupling}, this adaptive equilibrium demonstrates the superiority of our approach. 
Decoupling often suffered from a sharp decline in precision during the transition phase of image-text alignment. 
Additionally, we employed an alternating learning strategy, where vision and language prompts are trained for an equal number of iterations. 
In the alternating learning approach, vision and language prompts are trained in separate iterations, lacking synchronized integration between the vision and language modalities. 
In comparison, BBN reliance on training iterations falls short in dynamically balancing the weights for domain invariance and class separability, underscoring the proposed TPL's effectiveness. 
As the distances decreased, the Weighting mechanism dynamically shifted, reducing the emphasis on domain invariance and increasing the focus on class separability. 
This shift ensures that while the model benefits from the reduced domain distances, it does not compromise its ability to distinguish between different classes within a domain.

\section{Conclusion}
We propose a new vision-language prompt learning scheme (TPL) for the problem of DG by combining vision and language prompts through a transitive learning approach to address the balance between domain invariance and class separability. 
To ensure this balance, we propose an adaptive weighting strategy that dynamically calibrates the trading-off between domain invariance and class separability, and this trading-off can be seamlessly adapted to new and unseen domains. 
Experiments on different datasets show that the proposed TPL consistently outperforms existing generalization techniques.
The effectiveness of producing features with domain invariance and class separability is also validated through extensive experiments. 

{\small
\bibliographystyle{plainnat}
\bibliography{efficientsr}
}

\end{document}